\documentclass{article}
\usepackage{spconf,amsmath,graphicx}
\usepackage{multirow}
\usepackage{tabularx}
\usepackage{color, soul}
\usepackage{dblfloatfix}

\title{Hierarchical Attention Transformer Architecture for  Syntactic Spell Correction}
\name{Abhishek Niranjan, M Ali Basha Shaik, Kushal Verma}
\address{
  Voice Intelligence, Samsung Research - Bangalore, India\\
{\{a.niranjan, m.shaik, kushal.verma\}@samsung.com}}

\begin{document}
%
\maketitle
\begin{abstract}

  The attention mechanisms are playing a boosting role in advancements in sequence-to-sequence problems. Transformer architecture achieved new state of the art results in machine translation, and it's variants are since being introduced in several other sequence-to-sequence problems. Problems which involve a shared vocabulary, can benefit from the similar semantic and syntactic structure in the source and target sentences. With the motivation of building a reliable and fast post-processing textual module to assist all the text-related use cases in mobile phones, we take on the popular spell correction problem.   
  In this paper, we propose multi encoder-single decoder variation of conventional transformer. Outputs from the three encoders with character level 1-gram, 2-grams and 3-grams inputs are attended in heirarchical fashion in the decoder. The context vectors from the encoders clubbed with self-attention amplify the n-gram properties at the character level and helps in accurate decoding. 
  We demonstrate our model on spell correction dataset from Samsung Research, and report significant improvement of 0.11\%, 0.32\% and 0.69\% in characeter (CER), word (WER) and sentence (SER) error rates from existing state-of-the-art machine-translation architectures. Our architecture is also trains $\approx$7.8 times
faster, and is only about $\frac{1}{3}$ in size from the next most accurate model. 

\end{abstract} 

\vspace{-5mm}
\section{Introduction}
\label{sec:intro}

Quality of both, the transcriptions by ASR engines and the autocorrection in keyboards, have seen a major improvement in the recent years; introduction of deep neural architecture in acoustic and language modeling seems to be the major factor in speech recognition engines. Also the shift from n-gram based model to neural model in recent keyboards plays an important role in this. However, post processing of transcriptions generated by ASR systems or the correction wrongly user spelled sentences still has a huge scope of improvement. 

	Despite such advancements in speech recognition models, post hypothesis text-processing is still being dominated by rule based correction algorithms. Lack of sufficient labeled data, compute resources and high dependence over language modeling (in case of keyboard algorithms) adds to the difficulty of this problem. 
	
	Emergence of architectures, purely working over the concept of attention-mechanisms and removing the need of recurrent units, in machine translation has seen their successful involvement in other sequence-to-sequence problems as well, for e.g grapheme-to-phoneme, etc.
	
	We specifically focus on ASC task to further a step in on-device textual processing with the help of neural networks, to enable them tackle lighter text-to-text problems with sole dependence on self and cross attention mechanisms. In this paper, we propose a multi-encoder variant of vanilla transformer architecture, which feeds character level unigrams, bigrams and trigrams to the parallel encoders, with the only decoder attending these attention vectors in a heirarchical style. We compare the performance of our model with recent architetures in machine translation on the basis of error rates, training, decoding time and feasibility of on-device deployment. 
	
	The paper is organized as follows: section 2 details literature in the spelling correctoin research in ASR systems, and in general. In section 3, we describe the dataset used for evaluation. Network architecture is explained in section 4, with experiments detailed in section 5. Results are shared in section 6. Section 7 concludes the paper followed by future plans in section 8. Acknowledments are made under section 9. 

\vspace{-5mm}
\section{Prior Work}

Spell correction is now a core pillar in multiple technologies, like ASR systems, search engines, OCR engines, etc and thus have come a long way ahead of the noisy channel model variants~\cite{Shannon1948} ~\cite{Brill:2000:IEM:1075218.1075255} with statistical and rule based solutions. 
	
	~\cite{Errattahi2015AutomaticSR} presents an overview of previous works on error correction for ASR. Two of the highlighted research in this paper were by ~\cite{Sarma:2004:CSR:1613984.1614006} built an unsupervised model to detect and correct ASR errors using co-occurence analysis. ~\cite{Bassil2012ASRCE} demonstrated the improvement in error rates by post-editing ASR errors based on Microsoft N-Gram dataset ~\cite{Wang:2010:OMW:1855450.1855462}. 
	
	Deep learning solutions to spell correction began with the usage of character embeddings. ~\cite{E17-2027} used skip-gram network to generate embeddings for units comprising of only consonants or continuos vowels. They reported significant improvement on Birkbeck spelling error corpus\footnote{http://www.dcs.bbk.ac.uk/ ROGER/corpora.html}.    
	
	Besides the conventional statistical models, researchers have applied a number of complex sequence-to-sequence architectures to tackle automatic spell correction (ASC). ~\cite{Xie2016NeuralLC} adopted an encoder-decoder network with attention mechanism to work on Grammatical Error Correction (GEC) task. Most recent work ~\cite{DBLP:journals/corr/abs-1902-07178} proposed an external-language model to rescore the n-best hypothesis of end-to-end ASR system. The encoder-decoder architecture was demonstrated which employed multi-headed additive attention.  
	
	Besides ASR systems, there have been recent advances in the keyboard autocorrection results in mobile phones. A third party keyboard software SwiftKey\footnote{https://swiftkey.com/en/terms} emerged as the foremost keyboard system to entierly shift from n-gram based language model to neural network based. Grammarly\footnote{https://www.grammarly.com} has been the most popular spell and grammar correction tool while writing document online which suggests the requirement of such tool.  
	
	Seeking motivation from the previous work and potential of such a tool in mobile devices, we propose a neural architecture based spell correction module with the feasibiliy for on-device computation. Also our network is an improvement to the previous networks as we demonstrate a neural model blended with statistical features, i.e character level unigrams, bigrams and trigrams. Idea behind the same is to exploit the syntactic similarity in the source, target pairs as the vocabulary is shared. 

\vspace{-5mm}
\begin{table}[!htbp]
  \caption{Noisy generations for voice directed inputs}
  \label{tab:data_preprocess}
  \centering
  \begin{tabularx}{\linewidth}{l|l}
    \hline
    \textbf{phrase} & \textbf{noisy generation} 		\\
    \hline
    OLD \ul{\textit{RAILWAY}} STATION & OLD \ul{\textit{REAL}} \ul{\textit{WAY}} STATION \\  
    OLD RAILWAY \ul{\textit{STATION}} & OLD RAILWAY \ul{\textit{STASH}} \ul{\textit{IN}} \\            
    CATHOLIC \ul{\textit{CHURCH}} & CATHOLIC \ul{\textit{CHERCHE}} \\
    \ul{\textit{SECOND}} HALF & \ul{\textit{SICKENED}} HALF \\
    WORK \ul{\textit{BENCH}} & WORK \ul{\textit{BUNCH}} \\
    \hline
  \end{tabularx}
\end{table}

\vspace{-7mm}
\section{Dataset}
We used open-domain dataset, which is a subset of NOW corpus\footnote{https://www.english-corpora.org/now/}. Since our motivation is to build an on-device HAT ASC system, we aimed to work on shorter length sentences and hence extracted phrases from the corpus. We used TextRank python library\footnote{https://github.com/DerwenAI/pytextrank} ~\cite{PyTextRank} to extract phrases from the corpus. 

	To cater textual-processing for both voice enabled inputs as well as user typed sentences, we create noisy and erroneous phrases. We introduce upto 3-edit distances spelling errors per word to simulate wrongly typed sentence. For ASR systems, we use internal english lexicon to replace a word(or word sequence) with similar sounding one or multiple words when the phoneme sequences of original grapheme and replacing grapheme(sequence) have cosine similarity $>$ a thresholod \textit{t}, set to 0.6 after some trial and errors. 
	
	We randomly pick at most 5 noisy generations per phrase from the combined pool of edit-distance based and similar-sound based erroneous phrases.
	Table 1 shows a few similar sounding noisy generations which our algorithm produced.

\begin{table}[t]
  \caption{Dataset Description}
  \label{tab:data_info}
  \centering
  \begin{tabular}{cc}
	\hline    
    \#examples          &  4 million                                       \\
    maximum \#words in a phrase & 5                                 \\
    minimum \#words in a phrase & 2                          \\
    \hline
  \end{tabular}
\end{table}

\vspace{-5mm}
\begin{table}[!htbp]
\tiny
  \caption{Example source-target pair modification}
  \label{tab:data_preprocess}
  \centering
  \begin{tabular}{ccc}
	\hline    
    \textbf{source phrase} & \textbf{inputs to encoders} & \textbf{input to decoder}                										\\
	\hline    
    \multirow{5}{*}{SICKENED HALF}  & S  I  C  K  E  N  E  D  \#  H  A  L  F  &  \multirow{5}{*}{S E C O N D \# H A L F} 				\\      
    				  &	    & \\            
     				   & SI  IC  CK  KE  EN  NE  ED  \#  HA  AL  LF  &                                 \\
     				   &    & \\
     				   & SIC  ICK  CKE  KEN  ENE  NED  \#  HAL  ALF &                        			\\
    \hline
  \end{tabular}
\end{table}

\begin{figure*}[!htbp]
	\centering
  	\includegraphics[width=\textwidth, height=280pt]{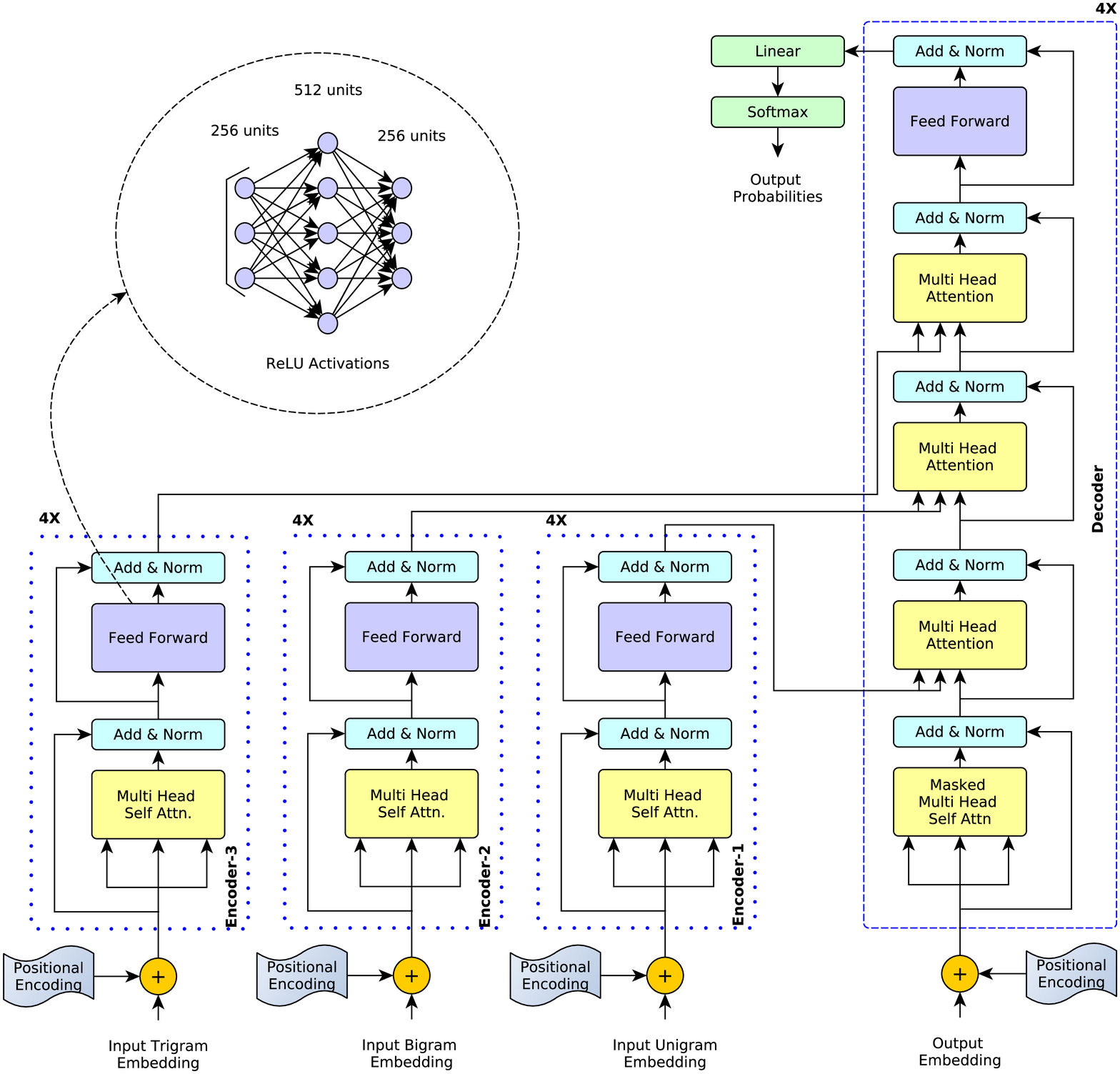}
  	\caption{Hierarchical Attention Transformer (HAT) Architecture block diagram}
  	\label{fig:arch}
\end{figure*}

\vspace{-6mm}
\subsection{Data Preprocessing}

We input character level unigrams, bigram and trigrams to the encoders of our architecture. Table 2 shows how we modify the raw source-target pair accordingly.

We also had to augment the dataset with true target-target phrase pairs for training and testing purposes. We injected such pairs to help the model distinguish the correctly spelled phrases from the wrong ones. The augmented target-target phrase pairs made 15\% of the whole dataset. Table 3 describes the final dataset  
	
	The data was distributed in 85\%, 5\% and 10\% split for training, dev and testing portions respectively.

\vspace{-5mm}
\section{Network Architecture}

Most sequence-to-sequence problems are now being solved with encoder-decoder type of architectures. Essentially the encoder transforms the input vector ($x_1, ..., x_n$) to a context vector \textbf{z} = ($z_1, ..., z_2$). This context vector is then fed to decoder, which autoregressively generates the output vector ($y_1, ..., y_2$) one element at a time.
The vanilla transformer follows the similar architecture. 

	We build over the fundamental architecture and introduce 3 encoders each feeding to the same decoder. \textbf{$z_1$}, \textbf{$z_2$} and \textbf{$z_3$} being the corresponding encoder outputs. Figure 1 details the proposed network architecture. 

\vspace{-5mm}
\subsection{Encoders and Decoder}
\textbf{Encoders: } We use 3 encoder stacks, one for each of the character level unigram inputs, bigram inputs and trigram inputs. All the encoder follow the exact same structure as ~\cite{Vaswani2017AttentionIA} only difference being the number of identical layers which is set to \textit{N} = 4, instead of 6 in the ~\cite{Vaswani2017AttentionIA}.
\\
\textbf{Decoder: }The decoder is different from \cite{Vaswani2017AttentionIA}. It replaces one encoder-decoder multi-head attention layer to three encoder-decoder multi-head attention layers, one for each of the encoders structured in a hierarchical fashion starting from encoder outputs \textbf{$z_1$} corresponding to unigram inputs and ending with encoder outputs \textbf{$z_3$} corresponding to trigram inputs.

\subsection{Attention mechanism}
\textbf{Attention Function}: We tried a combination of different functions ~\cite{D15-1166} in the attention layers in the encoders and decoder. Additive attention ~\cite{DBLP:journals/corr/BahdanauCB14}  and scaled dot-product attention ~\cite{Vaswani2017AttentionIA} performed nearly the same, we however ended up using scaled-dot attention as the model trained faster. 
Scaled Dot-Product Attention can be identified as:   \\ \\
 	\indent $Attention(Q, K, V ) = softmax(\frac{QK^{T}}{\sqrt{d_{k}}})V$ \\ \\
 The softmax of the dot product of the query $(Q)$ with all the keys $(K)$, scaled by a $\sqrt{d_{k}}$, where $d_{k}$ is the dimension of query and key vectors gives the weights(importance) to the keys, which is then multiplied by the values $(V)$ corresponding to the keys. \\ \\
\noindent\textbf{Multi-Head Attention and Attention Layers: } We follow the 
same idea of taking \textit{h} linear projections of the queries, keys and values, and apply atttention mechanism to each of these triplets ($q_{i}, k_{i}, v_{i}$).

	There are self-attention sub-layer at the base of every layer in encoders and decoder. Queries $(Q)$, keys $(K)$ and values $(V)$ in this case come from the same sequence and attends to the different positions of it's own sequence. The cross-attention layer is present only in the decoder stack where queries $(Q)$ come from the previous layer of decoder, and keys $(K)$ and values $(V)$ come from the encoder outputs. Three cross-attention layers are stacked in a heirarchical fashion in the sublayer of decoder, which attend to encoder outputs from unigram-level inputs, bigram-level inputs and trigram-level inputs in the same order.

	We keep the rest of model components same as the  \cite{Vaswani2017AttentionIA}.
\vspace{-5mm}
\subsection{Hyperparameters and Model Parameters}
We set the embedding size to 256, i.e the input and output tokens will be encoded into a 256 dimension vector. Number of hidden layers in encoder and decoder stacks, which are the combination of attention and feed-forward sublayers, is set to 4. We set the number of heads, number of linear projections of activations before undergoing attention mechanism to 8. The dimensionality of inner layer in feed-forward network is set to 512. 
	
	We employ the same 3 dropouts from ~\cite{Vaswani2017AttentionIA} and have these set to P\textsubscript{drop} = 0.3. We keep the rest of the hyperparameters same as ~\cite{Vaswani2017AttentionIA}.

\begin{table*}[htb]
	\caption{Error rates on dev and test data}
  	\label{tab:Results}
  	\centering
	\begin{tabular}[width=\textwidth]{lccc|ccc|c|c}
	\hline
	\textbf{Model} & \multicolumn{3}{c}{\textbf{dev}} & \multicolumn{3}{c}			{\textbf{test}} & \textbf{train time} & \textbf{model size\footnote{Uncompressed}} \\
	\hline
      & \textbf{CER}      & \textbf{WER}     & \textbf{SER}     & \textbf{CER}      		& \textbf{WER}      & \textbf{SER}  &   &  \\
	  	& (\%)    & (\%)    & (\%)    &(\%)	& (\%) & (\%) & (xRT) & (MBs) \\
  	\hline
     LSTM with attention~\cite{D15-1166} &  2.79 &  7.39 & 16.05 &  2.77 &  7.36  & 15.99  & 2.22 & \textbf{71}      \\
      DynamicConv ~\cite{wu2018pay}& 2.58 & 7.09  & 15.13 & 2.59  & 7.18          & 15.22  & 5.64  & 516 \\
      \hline
      Transformer - unigrams ~\cite{Vaswani2017AttentionIA}&    3.89& 10.89& 23.15        &     3.86 & 10.87  & 23.14 &  \textbf{0.46} & 98 \\
      + bigrams &  2.90& 8.01& 17.79 &   2.93 & 8.13  & 17.83 & 0.64 & 135\\
      + trigrams* &  \textbf{2.49} & \textbf{6.83} & \textbf{14.51} &  \textbf{2.48} & \textbf{6.86} & \textbf{14.53}  & 1.0 & 166 \\    
      \hline
      \multicolumn{9}{l}{\hspace{5pt}*real-time decoding for a 32 size batch completes in $\approx$10 ms  }
	\end{tabular}
\end{table*}

\vspace{-5mm}
\section{Experiments}
We started with estabilishing the baseline, i.e the vanilla transformer ~\cite{Vaswani2017AttentionIA}, results for this task.  We then proceeded to run the very recent neural machine translation architecture ~\cite{wu2018pay} which used lightweight and dynamic convolution kernels to work as a replacement of attention component in the fundamental transformer architecture. We trained the said model with 4 and 7 (default) hidden layers in the encoder and decoders and report the performances from $N$=7 network as it was more accurate than the former. 

	We then performed the experiment while taking character level unigram and bigram as inputs, and lastly stopped at three encoder based architecture as our aim was to produce a lightweight and accurate model which should have the flexibility to be able to compute on-device. 
	
	With a batch size of 2048 with at most 128 tokens per input sequence, we trained aforementioned models on 4 NVIDIA P40 GPUs for 30K steps.

\begin{table}[t]
  \caption{Additional Results on LibriSpeech(test-clean)}
  \label{tab:libri_results}
  \centering
  \begin{tabular}{ccc}
	\hline    
    Corpus          &  Duaration(hours) & WER(\%)                                          \\
    \hline
    DNN(p-norm)+SAT &     5.4 & 8.47                             \\
    S5(\textit{cf.} Table\ref{tab:Results}) & 5.4 & 7.80                          \\
    \hline
  \end{tabular}
  \vspace{-4mm}
\end{table}

\vspace{-4mm}
\section{Results}
\vspace{-3mm}
We compare our model's performance, size and training time with the following sequence-sequence architectures ~\cite{D15-1166}, ~\cite{wu2018pay} and ~\cite{Vaswani2017AttentionIA}. We obtain the lowest character (CER), word (WER) and sentence (SER) error rates in all the networks, an improvement of 0.11\%, 0.32\% and 0.69\% respectively from the next most accurate model and 1.38\%, 4.01\% and 8.61\% respectively from the vanilla transformer network.   
	 We also report better training time than ~\cite{wu2018pay}\footnote{Trained with 7 hidden layers} and ~\cite{D15-1166}. Our model size stands at 166 MBs, and is a trade-off between achieving high accuracy and the feasibility to be able to be deployed for on-device computation.
	Table 4 describes the results on our experiments.  
	
In addition, we also tested our proposed approach using Kaldi's standard s5 DNN recipe, trained on librispeech 100 hrs audio data. We achieved $\approx{0.67}$ improvment in WER on librispeech test-clean corpus as shown in Table \ref{tab:libri_results}.

\vspace{-5mm}
\section{Conclusions}
\vspace{-3mm}

In this paper we proposed a character n-gram driven heirarchical attention transformer variant to explicity solve the spell correction problem for voice and text relaed in mobile devices. We tried to exploit the shared vocabulary component of the problem and aimed to provide features to catch the syntactic similarity in source and target sentence pairs.
Our model show that passing unigrams, bigrams and trigrams as inputs to a shared vocabulary sequence-to-sequence problem yields clear improvement over the vanilla transformer architecture, and recent very successful networks in the neural machine translation domain. 
	We also discuss feasibility of model running on mobile devices as a post-processing step for text and voice enabled services due to it's advantages while training and decoding steps over traditional recurrent unit based encoder-decoder architectures ~\cite{D15-1166}. 


\vspace{-2mm}
\begin{figure}[!htbp]
	\centering
  	\includegraphics[width=\linewidth]{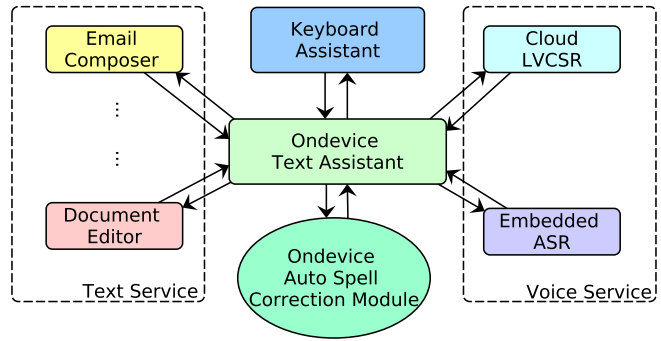}
  	\caption{Use-cases of lightweight textual-processing module in mobile devices}
  	\label{fig:arch}
  	\vspace{-5mm}
\end{figure}

\vspace{-5mm}
\section{Future work}
\vspace{-3mm}
We aim to base the results of this paper to build a deployable and personalized text-processing module for mobile phones. The motivation is to enable a highly accurate, fast and efficient post processing step which can be personalized according to the context and user's interaction with their smart phones. 

	We also look forward to better this network architecture for more prominent sequence-to-sequence tasks such as Grammatical Error Correction (GEC), Similar language translation, etc. where the source and target sentences share more or less a common vocabulary.

\vspace{-5mm}
\section{Acknowledgements}
\vspace{-3mm}
We gratefully acknowledge the support from Samsung Research for funding of the research project. We thank Chan Woo Kim and Dhananjaya Gowda for their encouragement and support.  
\bibliographystyle{IEEEbib}
\bibliography{strings,refs}

\end{document}